\documentclass[11pt,a4paper]{article}

\usepackage[final]{acl2023}

\usepackage{times}
\usepackage{latexsym}

\usepackage{amsmath}
\usepackage{amsfonts}
\usepackage[linesnumbered,ruled,vlined]{algorithm2e}
\usepackage{graphicx} 
\usepackage{booktabs}
\usepackage{array}
\usepackage{enumitem}
\usepackage{amsthm}
\usepackage{algpseudocode}
\usepackage[switch]{lineno}
\usepackage[utf8]{inputenc}
\usepackage{eqparbox}
\usepackage[nopar]{lipsum}
\usepackage{multirow}
\usepackage{makecell}
\usepackage{xcolor}
\usepackage{hhline} 
\usepackage{microtype}
\usepackage{diagbox}

\usepackage{adjustbox}

\usepackage{relsize}

\usepackage{booktabs,multirow,array}
\newcolumntype{N}{@{}m{0pt}@{}}

\usepackage[many]{tcolorbox}
\newtcolorbox{fancyquotes}{%
    enhanced jigsaw, 
    breakable,      
    frame hidden,   
    left=0.5cm,       
    right=0.1cm,      
    overlay={%
        \node [scale=8,
            text=black,
            inner sep=0pt,] at ([xshift=-1cm,yshift=-1cm]frame.north west){}; 
        \node [scale=8,
            text=black,
            inner sep=0pt,] at ([xshift=1cm]frame.south east){};  
            },
                parbox=false,
}

\usepackage{mathtools, nccmath}
\usepackage{scrextend}
\deffootnote[.25in]{.25in}{.15in}{\makebox[.25in][r]{\thefootnotemark .\hspace{.15in}}}

\makeatletter

\newtheorem*{proof*}{Proof}

\newcommand{\blue}[1]{\textcolor{blue}{#1}}

\newcommand{\green}[1]{\textcolor{green}{#1}}

\usepackage{listings}
\usepackage{color}
\definecolor{codegreen}{rgb}{0.3,0.5,0.0}
\def\@fnsymbol#1{\ensuremath{\ifcase#1\or \dagger\or *\or \ddagger\or
   \mathsection\or \mathparagraph\or \|\or **\or \dagger\dagger
   \or \ddagger\ddagger \else\@ctrerr\fi}}

\newcolumntype{C}[1]{>{\centering\let\newline\\\arraybackslash\hspace{0pt}}m{#1}}
\newcommand\ChangeRT[1]{\noalign{\hrule height #1}}

\newcommand\Vtextvisiblespace[1][.3em]{%
  \mbox{\kern.06em\vrule height.3ex}%
  \vbox{\hrule width#1}%
  \hbox{\vrule height.3ex}}

\newcolumntype{R}[2]{%
    >{\adjustbox{angle=#1,lap=\width-(#2)}\bgroup}%
    l%
    <{\egroup}%
}

\ExplSyntaxOn
\NewExpandableDocumentCommand { \ValuePlusOne } { m } 
  { \int_eval:n { \int_use:c { c @ #1 } + 1 } }
\NewExpandableDocumentCommand { \Sec } { m } 
  { \fp_eval:n { secd ( #1 ) } }
\NewDocumentCommand { \Rot } { m }
  { 
    \hbox_to_wd:nn { 1 em }
      { 
        \hbox_overlap_right:n 
          { 
            \skip_horizontal:n { \fp_to_dim:n { 7 * cosd (\Angle) } } 
            \rotatebox{\Angle}{#1}
          } 
      } 
  }
\ExplSyntaxOff

\def\Angle{45}
    
\bigskip
\def\Angle{90}

\title{Llama2Vec: Unsupervised Adaptation of Large Language Models for Dense Retrieval} 
\author{Zheng Liu$^{2,4}$\footnotemark[1], ~Chaofan Li$^{1}$\thanks{Co-first authors and major contributors of techniques}~\thanks{Major contributor of engineering}, ~ Shitao Xiao$^{2}$\footnotemark[1], ~ Yingxia Shao$^1$\thanks{Corresponding author.}, ~Defu Lian$^3$\\
  1: Beijing University of Posts and Telecommunications \\
  2: Beijing Academy of Artificial Intelligence \\
  3: University of Science and Technology of China \\
  4: The Hong Kong Polytechnic University \\
  \texttt{zhengliu1026@gmail.com} \quad
  \texttt{\{cfli, shaoyx\}@bupt.edu.cn} \\
  \texttt{stxiao@baai.ac.cn} \quad
  \texttt{liandefu@ustc.edu.cn}
}

\begin{document}
\maketitle

\begin{abstract}
Dense retrieval calls for discriminative embeddings to represent the semantic relationship between query and document. It may benefit from the using of large language models (LLMs), given LLMs' strong capability on semantic understanding. However, the LLMs are learned by auto-regression, whose working mechanism is completely different from representing whole text as one discriminative embedding. Thus, it is imperative to study how to adapt LLMs properly so that they can be effectively initialized as the backbone encoder for dense retrieval. 

In this paper, we propose a novel approach, called \textbf{Llama2Vec}, which performs unsupervised adaptation of LLM for its dense retrieval application. Llama2Vec consists of two pretext tasks: EBAE (Embedding-Based Auto-Encoding) and EBAR (Embedding-Based Auto-Regression), where the LLM is prompted to \textit{reconstruct the input sentence} and \textit{predict the next sentence} based on its text embeddings. Llama2Vec is simple, lightweight, but highly effective. It is used to adapt LLaMA-2-7B on the Wikipedia corpus. With a moderate steps of adaptation, it substantially improves the model's fine-tuned performances on a variety of dense retrieval benchmarks. Notably, it results in the new state-of-the-art performances on popular benchmarks, such as passage and document retrieval on MSMARCO, and zero-shot retrieval on BEIR. The model and source code will be made publicly available to facilitate the future research. Our model is available at \href{https://github.com/FlagOpen/FlagEmbedding}{https://github.com/FlagOpen/FlagEmbedding}.
\end{abstract} 


\section{Introduction}
Dense retrieval is a new paradigm of IR empowered by deep neural networks. It represents query and document as embeddings within the same latent space, where the semantic relationship between query and document can be reflected by their embedding similarity. Nowadays, dense retrieval has been a critical component in many real-world scenarios, like open-domain QA, fact verification, and retrieval-augmented generation \cite{karpukhin2020dense,thorne2018fever,lewis2020retrieval, Wang2022fact, Dolci2023}. 

The capacity of text encoder is a critical factor of dense retrieval. In the past few years, the pre-trained language models, e.g., BERT \cite{devlin2018bert}, RoBERTa \cite{Liu2019Roberta}, T5 \cite{raffel2020exploring}, were widely applied to generate high-quality representations for query and document, which substantially contributed to the accuracy and generalizability of dense retrieval. Besides, it was also found that dense retrieval's performance can further benefit from the continual growth of model size and training scale \cite{ni2021large,izacard2021unsupervised,wang2022text,xiao2023c}.  
Recently, large language models (LLMs) have emerged as a unified foundation for general NLP tasks \cite{brown2020language,wei2021finetuned,chowdhery2023palm}. Given the LLMs' superior capability on semantic understanding, it will be promising to take advantage of such powerful models as new backbones for dense retrieval. With this consideration, there have been pioneering efforts where LLMs are trained to generate discriminative text embeddings to facilitate the retrieval tasks in many different scenarios \cite{muennighoff2022sgpt,neelakantan2022text,ma2023fine,zhang2023language}. 

\begin{figure*}[t]
\centering
\includegraphics[width=1.0\textwidth]{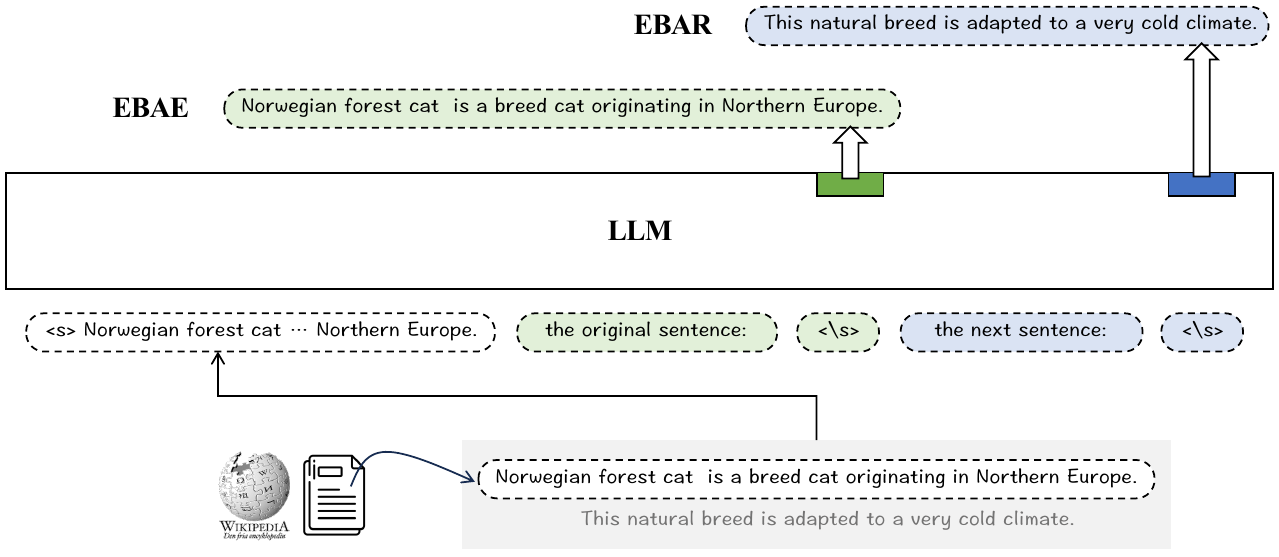}
\vspace{-15pt}
\caption{\textbf{Llama2Vec}. The LLM is prompted to generate the inductive embedding for EBAE (\green{green}), where the original sentence is predicted, and the deductive embedding for EBAR (\blue{blue}), where the next sentence is predicted.}  
\vspace{-5pt}
\label{fig:1}
\end{figure*}

Despite the preliminary progress, it remains an open problem to fully unleash the LLM's underlying potential for dense retrieval. Particularly, the typical LLMs are pre-trained by text generation tasks, especially auto-regression, where the text embeddings are learned to predict the next tokens. Consequently, the LLMs' text embeddings will focus on representing the local (i.e. the next step) semantic of the context. However, dense retrieval calls for text embeddings to represent the global semantic about the query and document. Such a big discrepancy will severely restrict the direct application of LLMs for dense retrieval. 

To address the above problem, we propose a novel approach \textbf{Llama2Vec} (Figure \ref{fig:1}), which performs unsupervised adaptation of LLMs to facilitate their usage in dense retrieval. Llama2Vec works as a continuation of pre-training. On top of the tailored unsupervised learning tasks, it transforms the LLM's text embeddings to represent the global semantic about the input text, which makes the LLM a better initialized encoder for dense retrieval. Llama2Vec is made up of two pretext training tasks: \textbf{EBAE} (embedding-based auto-encoding) and \textbf{EBAR} (embedding-based auto-regression). On top of EBAE, the LLM is prompted to generate the inductive text embedding, which can be used to predict the vocabularies of the input sentence itself. With EBAR, the LLM is prompted to generate the deductive text embedding, which can be used to predict the vocabularies for the next sentence. 

The joint conduct of the above pretext tasks bring forth two benefits. Firstly, the text embeddings from LLM can be adapted from \textit{local} semantic representation (predicting the next tokens) to \textit{global} semantic representation (predicting the sentence-level features), which aligns with the expected property of dense retrieval. Secondly, by learning to generate inductive and deductive text embeddings with different prompts, the LLM-based retriever can flexibly handle diversified semantic relationships about correlation (e.g., QA) and paraphrasing (e.g., NLI), which presents a strong foundation to develop versatile retrieval models. It's worth noting that the prediction is realized in the form of multi-classification, where the LLM's text embedding is the input and the vocabularies within the target sentence are employed as the labels. Therefore, Llama2Vec is extremely lightweight and simple to realize based on the existing auto-regression pipeline. 



We apply Llama2Vec for LLaMA-2-7B (base) \cite{touvron2023llama} over the Wikipedia corpus, where it substantially improves the LLM's downstream retrieval performance. With standard fine-tuning, the well-adapted model is able to notably outperform all existing dense retrieval methods, where it establishes new state-of-the-art performances on a variety of popular benchmarks, including the supervised tasks like passage and document retrieval of MSMARCO \cite{nguyen2016ms}, and the zero-shot retrieval of BEIR \cite{thakur2021beir}. 

To summarize, our work is highlighted by the following technical contributions. 1) We propose a new unsupervised learning method Llama2Vec. To the best of our knowledge, this is the first research work which explores the adaptation of LLMs for dense retrieval. 2) Llama2Vec is designed with simple but effective pretext tasks, which substantially improves the quality of LLM-based dense retriever in a cost-effective way. 3) The empirical studies verify the effectiveness of Llama2Vec, where substantial improvements can be achieved for both supervised and zero-shot retrieval tasks. 


\section{Related Works}

In this section, the related works are discussed from two perspectives: the background of dense retrieval, and the previous efforts on leveraging the LLMs for dense retrieval applications. 

$\bullet$ \textbf{Dense retrieval}. Dense retrieval is to represent query and document as embeddings within the same latent space, where relevant documents can be retrieved by embedding similarity. Nowadays, it is widely utilized in many important applications, such as open-domain QA and retrieval-augmented generation \cite{karpukhin2020dense,lewis2020retrieval}. The performance of dense retrieval is influenced by many factors. For example, dense retrieval models are learned by contrastive learning, where the discriminativeness of text embeddings are largely influenced by the scale and hardness of negative sample \cite{qu2020rocketqa,izacard2021unsupervised,xiong2020approximate}. Besides, the learning of dense retrieval models can benefit from knowledge distillation, where fine-grained teacher labels are derived from the ranking models \cite{hofstatter2021efficiently,ren2021rocketqav2}. Apart from the above training algorithms, the backbone architecture is one decisive factor for dense retrieval. In the past few years, pre-trained language models (PLMs), like BERT \cite{devlin2018bert}, RoBERTa \cite{Liu2019Roberta}, T5 \cite{raffel2020exploring}, have been widely adopted for the encoding of query and documents. Thanks to the large-scale model architecture and pre-training, PLMs were able to produce fine-grained semantic representation of input data, which substantially benefit the quality of dense retrieval. Besides, it was found that with the expansion of model and training scale, and the optimization of pre-training algorithm, the accuracy and generality of the PLM-based dense retrieval can be further improved.  \cite{ni2021large,izacard2021unsupervised,wang2022text,xiao2023c,gao2021condenser,liu2022retromae,liu2023retromae,wang2022simlm}. 

$\bullet$ \textbf{Dense retrieval with LLM}. The LLMs have been a unified foundation for many NLP tasks because of its superior capabilities. As a result, it is instinctive to leverage such powerful models to facilitate dense retrieval. LLMs can substantially contribute to many critical aspects of dense retrieval. For example, it can help to model the complex relationship between query and document considering LLMs' strong semantic understanding capability \cite{brown2020language,chowdhery2023palm,touvron2023llama}. Besides, it will benefit the learning of multi-task retrievers because of LLMs' versatility and instruction following capability \cite{wei2021finetuned,chung2022scaling}. It also presents a powerful foundation to develop long-document retrievers, given its dramatically extended context lengths. Recently, there have been several preliminary works which made important progresses on LLM-based dense retrieval \cite{muennighoff2022sgpt,neelakantan2022text,ma2023fine,zhang2023language}. However, the existing methods simply made direct use of LLMs. Because of the discrepancy between language modeling and text embedding, much of the LLMs' underlying potential is unexploited. In fact, it is still an open problem to study the proper adaptation of LLMs so that they can better contribute to dense retrieval.

\section{Llama2Vec} 

\subsection{Preliminary}
Dense retrieval utilizes a text embedding model to produce the query and document's embedding: $\boldsymbol{e}_q$ and $\boldsymbol{e}_d$. The relevance of query and document is reflected by their embedding similarity: $\langle \boldsymbol{e}_q, \boldsymbol{e}_d \rangle$. As such, the relevant documents for the query ($D_q$) can be retrieved via the ANN search within the embedding space: $D_q \leftarrow \text{Top-}k(\{d: \langle \boldsymbol{e}_q, \boldsymbol{e}_d \rangle | D\})$. 

The pre-trained language models used to be the backbone architecture of the embedding model. Take BERT as an example. The input text is tokenized as the sequence $T$: [CLS], t1, ..., tN, [EOS]. Then, the tokenized sequence is encoded by BERT, where the output embeddings are integrated as the text embedding. There are two common options to perform the integration: [CLS], or mean-pooling: 
\begin{align*}
    & \boldsymbol{e}_t \leftarrow \text{BERT}(T)[\text{CLS}] ~\textit{or}~ \text{AVG}\big(\text{BERT}(T)\big). 
\end{align*}
When using LLMs as the encoding architecture, the text embedding needs to be generated in a different way. Since the existing LLMs mainly use the decoder-only architecture \cite{brown2020language,chowdhery2023palm,touvron2023llama}, the global context can only be accessed by the very last tokens of the input sequence. Therefore, the output embedding from the special token $\langle\text{\textbackslash{s}}\rangle$ or [EOS] is utilized to represent the input text \cite{zhang2023language,ma2023fine}. Taking LLaMA \cite{touvron2023llama} as the example, we have the following updated form of text embedding: 
\begin{equation}
    \boldsymbol{e}_t \leftarrow \text{LLaMA}(T)[\langle\text{\textbackslash{s}}\rangle]. 
\end{equation} 

\begin{figure}[t]
\centering
\includegraphics[width=0.75\linewidth]{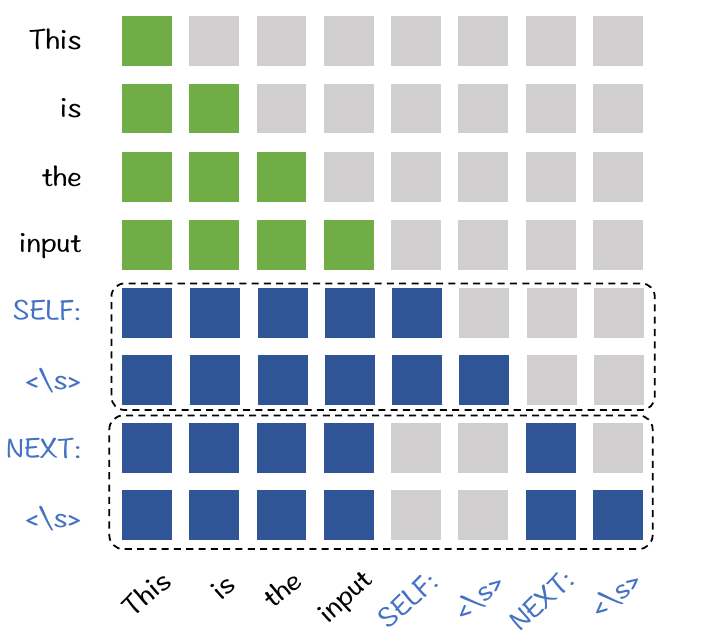}
\vspace{-5pt}
\caption{The attention scheme of Llama2Vec.}   
\vspace{-10pt}
\label{fig:2}
\end{figure}

\subsection{Unsupervised Adaptation}\label{sec:llara}
As introduced, the LLM's output embedding tends to foucs on local semantic, that is the prediction of the next token, because of its language modeling-based pre-training. To facilitate the application in dense retrieval, we perform unsupervised adaptation of LLM, where the LLM's text embedding can be transformed into a semantic representation of the global context.  

$\bullet$ \textbf{Objective}. The retrieval tasks can be roughly divided into two types. One is to find the correlated data, e.g., question-answering (QA). The other one is to identify paraphrasing data, e.g., natural language inference (NLI). To confront these tasks, text embeddings are expected to fulfill two objectives: 
\begin{itemize}[leftmargin=15pt] 
    \item \textit{Induction}. The representation of global semantic about the input text itself.
    \item \textit{Deduction}. The representation of global semantic for the correlated text of the input. 
\end{itemize}

$\bullet$ \textbf{Pretext tasks}. With the above objectives, we present two pretext training tasks of Llama2Vec. One is called \textbf{EBAE} (Embedding-based Auto-Encoding), where the text embedding $\boldsymbol{e}_t$ is used to induct the sentence-level feature about input sentence itself. 
The other one is called \textbf{EBAR} (Embedding-based Auto-Regression), where the text embedding $\boldsymbol{e}_t$ is used to deduct the sentence-level feature for the next sentence of the input. 
We argue that the strong induction and deduction capability of text embedding will be sufficient to handle the diversified retrieval scenarios, considering that arbitrary correlations can always be abstracted into the general form of \textit{input} (e.g., question), \textit{retrieval purpose} (e.g., get its answer), \textit{next sentence} (e.g, answer). 


$\bullet$ \textbf{Text Embedding}. The LLM is prompted by two different templates to generate the text embeddings for EBAE and EBAR (Figure \ref{fig:1}). For EBAE, the LLM is prompted by: ``[Placeholder for input]\Vtextvisiblespace[0.8em]{SELF}\Vtextvisiblespace[0.8em]$\langle\text{\textbackslash{s}}\rangle$'', where the inductive text embedding is generated by the following function: 
\begin{equation}
    \boldsymbol{e}^{\alpha}_t \leftarrow \text{LLaMA}(T,\text{SELF},\langle\text{\textbackslash{s}}\rangle)[-1].
\end{equation} 
SELF stands for the prompt of EBAE: ``The input sentence is:''. For EBAR, the LLM is prompted by template: ``[Placeholder for input]\Vtextvisiblespace[0.8em]{NEXT}\Vtextvisiblespace[0.8em]$\langle\text{\textbackslash{s}}\rangle$'', based on which the deductive text embedding is generated by the following function: 
\begin{equation}
    \boldsymbol{e}^{\beta}_t \leftarrow \text{LLaMA}(T,\text{NEXT},\langle\text{\textbackslash{s}}\rangle)[-1],
\end{equation} 
In this place, NEXT stands for the prompt of EBAR: ``The next sentence is:''. 

The direct computation of the two embeddings will lead to substantial unnecessary costs because the input text $T$ is repetitively processed for two times. To alleviate this problem, we propose to compute $\boldsymbol{e}^{\alpha}_t$ and $\boldsymbol{e}^{\beta}_t$ in one pass. Particularly, the prompts of EBAE and EBAR are merged into one joint prompt for LLM: ``[Placeholder for input]\Vtextvisiblespace[0.8em]{SELF}\Vtextvisiblespace[0.8em]$\langle\text{\textbackslash{s}}\rangle$\Vtextvisiblespace[0.8em]{NEXT}\Vtextvisiblespace[0.8em]$\langle\text{\textbackslash{s}}\rangle$''. Because the two text embeddings need to be computed independently, we modify the typical attention mask of casual language modeling, where 
``{SELF}\Vtextvisiblespace[0.8em]$\langle\text{\textbackslash{s}}\rangle$'' and ``{NEXT}\Vtextvisiblespace[0.8em]$\langle\text{\textbackslash{s}}\rangle$'' are mutually invisible (Figure \ref{fig:2}). Now, the output embeddings of the first and second $\langle\text{\textbackslash{s}}\rangle$ tokens are used for $\boldsymbol{e}^{\alpha}_t$ and $\boldsymbol{e}^{\beta}_t$, respectively. Given that the input text $T$ account for the majority of length for the joint prompt, such an operation will save almost 50\% of the computation cost compared with the naive method.

\renewcommand{\arraystretch}{1.2}
\begin{table*}[t]
    \centering
    \small
    \begin{tabular}{p{5.6cm}|C{1.0cm}|C{1.0cm}|C{1.2cm}|C{1.2cm}|C{1.2cm}|C{1.2cm}}
    \ChangeRT{1pt} 
    \multicolumn{3}{c}{-} & \multicolumn{2}{|c|}{\textbf{Dev}} & \textbf{DL'19} & \textbf{DL'20} \\
    \midrule
    \textbf{Method} & \textbf{Size} & \textbf{FT.} &
    \textbf{M@10} & \textbf{R@1000} & \textbf{N@10} & \textbf{N@10} \\ 
    \midrule
    BM25 \cite{lin2021pyserini} & -- & -- & 18.4 & 85.3 & 50.6 & 48.0 \\
    ANCE \cite{xiong2020approximate}  & 125M & hard & 33.0 & 95.9 & 64.8 & -- \\
    ADORE \cite{zhan2021optimizing}  & 110M & hard & 34.7 & -- & 68.3 & -- \\
    Condenser \cite{gao2021condenser}  & 110M & hard & 36.6 & 97.4 & 69.8 & -- \\
    coCondenser \cite{gao2021unsupervised}  & 110M & hard & 38.2 & 98.4 & 71.7 & 68.4 \\
    TAS-B \cite{hofstatter2021efficiently}  & 55M & distill & 34.3 & 97.6 & 72.2 & 69.2 \\
    RocketQAv2 \cite{ren2021rocketqav2}  & -- & distill & 38.8 & 98.1 & -- & -- \\
    AR2+SimANS \cite{zhou2022simans}  & 110M & distill & 40.9 & 98.7 & -- & -- \\
    GTR-XXL \cite{ni2021large} & 4.8B & -- & 38.8 & 99.0 & -- & -- \\ 
    SimLM \cite{wang2022simlm}   & 110M & hard & 39.1 & 98.6 & 69.8 & 69.2 \\
    SimLM+distill \cite{wang2022simlm}   & 110M & distill & 41.1 & 98.7 & 71.4 & 69.7 \\
    RetroMAE \cite{liu2022retromae} & 110M & hard & 39.3 & 98.5 & -- & -- \\
    RetroMAE+distill \cite{liu2022retromae} & 110M & distill & 41.6 & 98.8 & 68.1 & -- \\
    RetroMAEv2+distll \cite{liu2023retromae} & 110M & distill & 42.6 & 98.9 & \textbf{75.1} & -- \\
    LLaMA2-RepLLaMA \cite{ma2023fine}  & 7B & hard & 41.2 & 99.4 & 74.3 & 72.1\\
    OpenAI-Ada-002 \cite{neelakantan2022text}  & -- & -- & 34.4 & 98.6 & 70.4 & 67.6 \\
    \midrule
    LLaMA2-Llama2Vec & 7B & hard & \textbf{43.1} & \textbf{99.5} & 73.4 & \textbf{72.9} 
    \\
    \ChangeRT{1pt}
    \end{tabular}
    \vspace{-5pt}
    \caption{MS MARCO passage retrieval (performance measured by MRR@10, Recall@1000, NDCG@10).} 
    \vspace{-10pt}
    \label{tab:1}
\end{table*} 

$\bullet$ \textbf{Training}. As introduced, the text embeddings of Llama2Vec are adapted to capture the global semantic of the input sentence itself and the next sentence of the input. In this place, we propose a simple but effective training objective to accomplish such an adaptation. We argue that \textit{if one embedding is able to accurately predict the vocabularies in a specific context all by itself}, the embedding must be a strong representation of the global semantic for the corresponding context. 
Based on this fundamental principle, the training of text embedding is formulated as a multi-classification problem. It linearly projects the text embedding into the vocabulary space, where the vocabulary IDs of all tokens within the target context are predicted. Specifically, the objective function of this problem is derived as: 
\begin{equation}
    \min. - \frac{1}{|T|} \sum_{t \in \mathcal{T}} log \frac{\exp{(\boldsymbol{e}^T\boldsymbol{W}_{t}})}{\sum_{v \in V} \exp{(\boldsymbol{e}^T\boldsymbol{W}_{v}})}.
\end{equation}
In this place, $\boldsymbol{W} \in \mathbb{R}^{d \times |V|}$ is the projection head of LLM; $V$ indicates the vocabulary space; $\mathcal{T}$ stands for the collection of tokens of the target context (input text itself for $\boldsymbol{e}^{\alpha}_t$, the next sentence for $\boldsymbol{e}^{\beta}_t$). The above training objective is lightweight and simple to realize, which can be directly conducted based on the typical language modeling pipeline. 

$\bullet$ \textbf{Fine-Tuning}. The well-adapted LLM from Llama2Vec is fine-tuned for dense retrieval applications through contrastive learning. Because the majority of the fine-tuning datasets for dense retrieval are collected for correlation scenarios, such as QA \cite{nguyen2016ms} and Natural Question \cite{kwiatkowski2019natural}, which are made up of tuples of (query, answer), we can derive the following general form of objective function: 
\begin{equation}
    \min \sum_{q} - \log \frac{\exp(\langle \boldsymbol{e}_q^{\alpha},\boldsymbol{e}_a^{\beta} \rangle)}{\sum_{a' \in \mathcal{A}'} \exp(\langle \boldsymbol{e}_q^{\alpha},\boldsymbol{e}_{a'}^{\beta} \rangle)},
\end{equation}
where $\boldsymbol{e}_q^{\alpha}$ is the query's embedding prompted by NEXT, and $\boldsymbol{e}_a^{\beta}$ is the answer's embedding prompted by SELF. Despite the fixed formulation during training, the prompt scheme can be flexibility adjusted for each individual downstream scenario. Particularly, when dealing with the correlation relationships, e.g., question-answering, we hold on to NEXT and SELF to prompt the query and answer's embeddings. However, when handling other situations about paraphrasing relationships, the prompt scheme is as follows. To analyze the paraphrasing relationship between two long documents, we use SELF to prompt the query and answer's embeddings given its nature of summarizing the semantic of complex input. Meanwhile, we employ NEXT as the prompt for both inputs when dealing with two short sentences because of its nature of deducting the semantic for the related texts. 

\begin{table*}[t]
    \centering
    \small
    \begin{tabular}{p{5.0cm}|C{0.8cm}|C{0.8cm}|C{1.6cm}|C{1.5cm}|C{1.6cm}|C{1.6cm} }
    \ChangeRT{1pt} 
    \multicolumn{3}{c}{-} & \multicolumn{2}{|c|}{\textbf{Dev}} & \textbf{DL'19} & \textbf{DL'20} \\
    \midrule 
    \textbf{Method} & \textbf{Size} & \textbf{FT.} &
    \textbf{MRR@100} & \textbf{R@100} & \textbf{NDCG@10} & \textbf{NDCG@10} \\ 
    \hline
    BM25 \cite{lin2021pyserini} & -- & -- & 27.7 & 80.9 & 51.9 & 52.9 \\ 
    PROP \cite{ma2021prop} & 110M & -- & 39.4 & 88.4 & 59.6 & -- \\
    B-PROP \cite{ma2021b} & 110M & -- & 39.5 & 88.3 & 60.1 & -- \\
    COIL \cite{gao2021coil} & 110M & hard & 39.7 & -- & 63.6 & -- \\
    ANCE (first-p) \cite{xiong2020approximate} & 125M & hard & 37.7 & 89.3 & 61.5 & -- \\
    ANCE (max-p) \cite{xiong2020approximate} & 125M & hard & 38.4 & 90.6 & 62.8 & -- \\
    ADORE \cite{zhan2021optimizing} & 110M & hard & 40.5 & 91.9 & 62.8 & -- \\
    COSTA \cite{ma2022pre} & 110M & hard & 42.2 & 91.9 & 62.6 & -- \\
    LLaMA2-RepLLaMA \cite{ma2023fine} & 7B & hard & 45.6 & -- & 65.0 & 63.2 \\
    \hline
    LLaMA2-Llama2Vec & 7B & hard & \textbf{47.9} & \textbf{94.1} & \textbf{68.2} & \textbf{63.6} 
    \\
    \ChangeRT{1pt}
    \end{tabular}
    \vspace{-5pt}
    \caption{MS MARCO document retrieval} 
    \vspace{-5pt}
    \label{tab:2}
\end{table*} 

\section{Experiment} 
\subsection{Settings}
The experimental study is performed to explore three important issues: 1) Llama2Vec's retrieval performance after fine-tuning, 2) Llama2Vec's generalization across diversified scenarios, 3) the impact of technical factors in Llama2Vec. With such objectives, we use the MS MARCO \cite{nguyen2016ms} as our fine-tuning dataset, and perform the evaluation on the passage retrieval and document retrieval task. We also take advantage of the BEIR benchmark \cite{thakur2021beir}, where the fine-tuned retriever from MS MARCO is evaluated under the zero-shot setting to analyze its generalization capability. 

$\bullet$ \textbf{Training}. Llama2Vec is applied to the LLaMA-2-7B (base) model. The unsupervised adaptation is performed based on the unlabeled corpus of Wikipedia curated by DPR \cite{karpukhin2020dense}. We perform 10,000 steps of Llama2Vec adaptation in total, with a batch size of 256, a sequence length of 1024, and a learning rate of 1e-5. Llama2Vec is fine-tuned based on the training recipe presented by RepLLaMA \cite{ma2023fine}: it leverages LoRA \cite{hu2021lora} for the parameter efficient training of LLM, and simply relies on the ANN hard negatives \cite{xiong2020approximate} to fine-tune the embedding model with contrastive learning. 

\vspace{-2pt}
\subsection{Supervised Performance}
First of all, we analyze the supervised retrieval quality of Llama2Vec, where the model is fine-tuned with the training queries from MS MARCO passage and document retrieval, respectively. 

$\bullet$ \textbf{Passage Retrieval}. The experiment results on MS MARCO passage retrieval are shown in Table \ref{tab:1}. We make comparison with a wide variety of baseline methods on passage retrieval, which include the following categories: 1) basic lexical retriever: BM25 \cite{lin2021pyserini}; 2) dense retrievers fine-tuned from BERT or RoBERTa: ANCE \cite{xiong2020approximate}, ADORE \cite{zhan2021optimizing}, AR2+SimANS \cite{zhou2022simans}, RocketQAv2 \cite{ren2021rocketqav2}; 3) dense retrievers fine-tuned from the enhanced PLMs: Condenser \cite{gao2021condenser}, coCondenser \cite{gao2021unsupervised}, RetroMAE \cite{liu2022retromae,liu2023retromae}, SimLM \cite{wang2022simlm}; 4) dense retrievers based on LLMs: GTR-XXL based on T5-4.8B \cite{ni2021large}, SGPT \cite{muennighoff2022sgpt} and OpenAI-Ada-002 \cite{neelakantan2022text} based on GPT, RepLLaMA \cite{ma2023fine} based on LLaMA-2-7B. RepLLaMA is the closest baseline to our method, which directly fine-tunes the original LLaMA-2-7B backbone without any adaptation. There are two different fine-tuning methods (FT.): one is based on hard-negative sampling (hard): which is simple and low-cost; the other one is based on knowledge distillation (distill), which is accurate but expensive due to its demand of a precise ranker and complicated training process.  

\begin{table*}[t]
    \centering
    \small
    \begin{tabular}{p{1.5cm}|C{1.2cm}|C{1.2cm}|C{1.4cm}|C{1.4cm}|C{1.2cm}|C{1.2cm}|C{1.6cm}|C{1.6cm}}
    \ChangeRT{1pt}
    \textbf{Method} & BM25 & BERT & GTR-XXL & CPT-XL & Ada-2 & SGPT & RepLLaMA & Llama2Vec \\
    \hline
    \textbf{Size} & -- & 110M  & 4.8B  & 175B & --  & 5.8B  & 7B  & 7B \\
    \hline
    T-COVID & 59.5 & 61.5  & 50.1  & 64.9  & 81.3  & \textbf{87.3}  & 84.7  & \underline{86.9} \\
    NFCorpus & 32.2 & 26.0  & 34.2  & \textbf{40.7}  & 35.8  & 36.2  & 37.8  & \underline{38.2} \\
    NQ & 30.6 & 46.7  & 56.8 & -- & 48.2  & 52.4  & \underline{62.4}  & \textbf{64.6} \\
    HotpotQA & 63.3 & 48.8  & 59.9  & \underline{68.8} & 65.4  & 59.3  & 68.5  & \textbf{70.1} \\
    FiQA & 23.6 & 25.2  & 46.7  & \textbf{51.2} & 41.1  & 37.2  & 45.8  & \underline{48.5} \\
    ArguAna & 39.7 & 26.5  & 54.0  & 43.5  & \textbf{56.7}  & 51.4  & 48.6  & \underline{56.5} \\
    Touche & \textbf{44.2} & 25.9  & 25.6  & 29.1  & 28.0  & 25.4  & 30.5  & \underline{34.2}\\
    Quora & 78.9 & 78.7  & \textbf{89.2}  & 63.8  & 87.6  & 84.6  & 86.8  & \underline{88.3} \\
    DBPedia & 31.8 & 31.4 & 40.8  & \underline{43.2}  & 40.2  & 39.9  & 43.7  & \textbf{45.9} \\
    SCIDOCS & 14.1 & 11.3  & 16.1  & --  & 18.6  & \textbf{19.7}  & 18.1  & \underline{18.9} \\ 
    FEVER & 65.1 & 68.2 & 74.0  & 77.5  & 77.3  & 78.3  & \textbf{83.4}  & \underline{81.3} \\ 
    C-FEVER & 16.5 & 18.7   & 26.7  & 22.3  & 23.7  & 30.5  & \underline{31.0}  & \textbf{38.2} \\ 
    SciFact & 67.9 & 53.3 & 66.2  & \underline{75.4}  & 73.6  & 74.7  & \textbf{75.6}  & {74.8} \\ 
    CQA & 32.5 & 28.2 & 39.9 & - & \underline{41.7} & 38.1 & 37.4 & \textbf{43.2} \\
    \hline
    AVERAGE & 42.9 & 39.3  & 48.6  & --  & 51.4  & 52.1  & \underline{53.9} & \textbf{56.4} \\ 
    \ChangeRT{1pt}
    \end{tabular}
    \vspace{-5pt}
    \caption{Zero-shot retrieval on BEIR benchmark. (The performances are measured by NDCG@10)} 
    \vspace{-5pt}
    \label{tab:3}
\end{table*} 

\begin{table}[t]
    \centering
    \small
    \begin{tabular}{p{2cm}|C{1.8cm}|C{1.8cm}} 
    \ChangeRT{1pt}
    \textbf{Method} & \textbf{MRR} & \textbf{Hit Rate} \\  
    \hline
    bge-m3 & 69.67 & 88.94 \\
    OpenAI-TE3-S  & 65.69 & 89.42 \\
    OpenAI-TE3-L  & 67.37 & 90.38 \\
    RepLLaMA  & \underline{68.43} & \underline{91.35} \\
    Llama2Vec  & \textbf{70.56} & \textbf{92.31} \\ 
    \ChangeRT{1pt}
    \end{tabular}
    \vspace{-5pt}
    \caption{Zero-shot retrieval on Llama Index.} 
    \vspace{-10pt}
    \label{tab:llama-index}
\end{table}

The primary observations are presented as follows. First of all, Llama2Vec achieves a superior retrieval performance in every evaluation scenario. Remarkably, it achieves a MRR@10 of \textbf{43.1} and a Recall@1000 of \textbf{99.5}, which notably improves the performance of the baselines and presents a new state-of-the-art result on the large-scale dev set. Its performance is also highly competitive on DL'19 and DL'20, though it's slightly lower on DL'19 due to the randomness of the small test set. Besides, it leads to a notably improvement over the closest baseline RepLLaMA (based on the same backbone but without adaptation), which indicates the effect introduced by the adaptation of Llama2Vec. Finally, we can observe the the LLM-based retrievers' overwhelming advantages in comparison with the previous ones based on smaller PLMs, despite that the they are usually fine-tuned with a relatively simple approach (hard). Compared with the best PLM baseline fine-tuned by hard negatives, RetroMAE+hard and SimLM+hard, the switch to Llama2Vec brings forth almost +4\% gains in MRR@10. Such a dramatic improvement validates the LLMs' huge potential for dense retrieval, and with proper adaptation, such a potential can be exploited more effectively. 

$\bullet$ \textbf{Document Retrieval}. We report the evaluation results on MS MARCO document retrieval in Table \ref{tab:2}. We make comparison with popular document retrieval methods, including BM25 \cite{lin2021pyserini}, ADORE \cite{zhan2021optimizing}, ANCE first-p and max-p \cite{xiong2020approximate}, PROP \cite{ma2021prop}, B-PROP \cite{ma2021b}, COIL \cite{gao2021coil}, COSTA \cite{ma2022pre}, RepLLaMA \cite{ma2023fine}, which fall into the same categories as the passage retrievers.  

Our observations on document retrieval is very similar with our previous result on passage retrieval. In particular, Llama2Vec achieves a superior empirical performance in every evaluation, where it notably improves the 
previous BERT-based methods by \textbf{+5.7\%} point in MRR@100. Both LLM-based retrievers, RepLLaMA and Llama2Vec, are able to dominate the PLM-based baselines. Besides, Llama2Vec continues to outperform RepLLaMA, with a \textbf{+2.3\%} improvement in MRR@100 on the large-scale dev set and consistent advantages on DL'19 and DL'20. The above observations further affirm our previous conclusions about the advantage of LLM backbone and the benefit from Llama2Vec. It's worth noting that the LLM presents a powerful backbone to support document retrieval, because of not only its high expressiveness but also its long context, which enables the input document to be fully encoded instead of chunked into smaller segments. 

\renewcommand{\arraystretch}{1.4}
\begin{table*}[t]
    \centering
    \small
    \begin{tabular}{p{0.7cm}|C{0.56cm}|C{0.56cm}|C{0.56cm}|C{0.56cm}|C{0.56cm}|C{0.56cm}|C{0.56cm}|C{0.56cm}|C{0.56cm}|C{0.56cm}|C{0.56cm}|C{0.56cm}|C{0.56cm}|C{0.56cm}|C{0.56cm}} 
    \ChangeRT{1pt}
    \textbf{Pro.} & \textbf{Avg} & \textbf{AR}& \textbf{CF} & \textbf{DB} & \textbf{FV} & \textbf{FQ} & \textbf{HQ} & \textbf{NF} & \textbf{NQ} & \textbf{QR} & \textbf{SD} & \textbf{SF} & \textbf{TO} & \textbf{TC} &
    \textbf{CQ}  \\
    \ChangeRT{1pt}
    \textbf{N2S} & 56.1 & 50.5 & 27.7 & 45.9 & 83.5 & 48.5 & 70.1 & 38.2 & 64.6 & 83.7 & 18.9 & 76.5 & 34.2 & 86.9 & 41.2 \\
    \textbf{S2S} & 30.2 & 56.5 & 20.6 & 10.4 & 33.1 & 16.7 & 27.4 & 18.3 & 8.8 & 88.0 & 7.1 & 67.8 & 3.1 & 34.2 & 26.0 \\
    \textbf{N2N} & 52.3 & 49.9 & 38.2 & 40.5 & 81.3 & 43.8 & 65.5 & 35.3 & 54.9 & 88.3 & 20.5 & 74.8 & 22.6 & 63.9 & 43.2 \\
    \textbf{None} & 47.6 & 53.8 & 26.7 & 40.0 & 72.3 & 38.9 & 63.8 & 32.3 & 46.1 & 88.3 & 17.7 & 74.3 & 15.0 & 49.6 & 41.1 \\
    \textbf{Ada*} & 57.4 & 56.5 & 38.2 & 45.9 & 81.3 & 48.5 & 70.1 & 38.2 & 64.6 & 88.3 & 18.9 & 74.8 & 34.2 & 86.9 & 43.2 \\ 
    \ChangeRT{1pt}
    \end{tabular}
    \vspace{-5pt}
    \caption{Impact of the adaptive usage of prompt (Ada*) evaluated on BEIR.} 
    \vspace{-10pt}
    \label{tab:4}
\end{table*} 

\subsection{Zero-shot Performance} 
We further investigate Llama2Vec's impact on the generalization. We leverage the BEIR benchmark and Llama Index for the evaluation of zero-shot performances, where the fine-tuned model from MS MARCO is directly applied to its included datasets (Table \ref{tab:3} and Table \ref{tab:llama-index}). 

\begin{table}[t]
    \centering
    \small
    \begin{tabular}{p{1.2cm}|C{1.4cm}|C{1.4cm}|C{1.4cm}} 
    \ChangeRT{1pt}
    \textbf{Top-N} & \textbf{Initial} & \textbf{Adapt} & \textbf{Fine-Tune} \\  
    \hline
    10   & 1.85 & 2.74 & 13.83 \\
    100  & 18.01 & 47.68 & 84.08 \\
    500  & 93.68 & 205.42 & 307.30 \\
    1000  & 219.09 & 392.80 & 542.89 \\ 
    \ChangeRT{1pt}
    \end{tabular}
    \vspace{-5pt}
    \caption{Impact on lexical similarity.} 
    \vspace{-10pt}
    \label{tab:5}
\end{table}

The major observations about the BEIR evaluation result are presented as follows. First of all, Llama2Vec exhibits a remarkable performance on BEIR, where it achieves a average performance of \textbf{56.4}. Such a performance is not only much higher than the rest of the baselines, but also establishes a new state-of-the-art result on BEIR (zero-shot). More impressively, it maintains the leading (bold) or the 2nd-place performance in almost every dataset, which indicates its superior versatility across different scenarios. Besides, Llama2Vec substantially outperforms RepLLaMA in most of the scenarios (12/14), which indicates the comprehensive improvement of the retriever's generalization. It is also worth to emphasize the comparison between BM25 and dense retrievers. When the BEIR benchmark was first launched two years ago, none of the dense retrievers (BERT and many others) were able to outperform BM25 despite their competitive performances in the supervised scenarios. However, the previous situation has been largely overturned with the adoption of LLM-based text encoders, as Llama2Vec outperforms BM25 on 13/14 datasets and goes beyond its average performance by 31\% relatively. The dramatic improvement can attribute to three merits of LLMs: 1) the superior expressiveness to model complex semantics, 2) the expanded context to handle long inputs, and 3) the rich knowledge to understand common-sense relationships. 

Meanwhile, Llama2Vec exhibits exceptional performance on the Llama Index. Notably, Llama2Vec surpasses bge-m3 \cite{chen2024bge}, OpenAI's text-embedding-3 in both small and large embedding sizes, indicating that Llama2Vec is highly competitive compared to the universal models in the community. When compared to RepLLaMa, which is only fine-tuned on the MS MARCO passage dataset, Llama2Vec also demonstrates stronger generalization abilities, showing better performance in both MRR and Hit Rate.

\subsection{Technical Factors}
Further study is made for three factors: unsupervised adaptation, prompt scheme, embedding size. 

$\bullet$ \textbf{Adaptation}. Our previous experiments verify the effectiveness of unsupervised adaptation given its substantial improvement over LLaMA-2-7B. In this place, we focus on exploring the the underlying reason of the empirical advantage. As introduced, the unsupervised adaptation is performed to transform the text embedding such that it can become a global semantic representation and make accurate predictions for the vocabularies within the target context (\S \ref{sec:llara}). We perform the following experiment on MS MARCO to evaluate the adaptation effect.
Firstly,  the query and answer embeddings are projected into distributions in the vocabulary space: $\boldsymbol{d}_q \in \mathbb{R}^{|V|\times1} \leftarrow {\boldsymbol{e}_q^{\alpha}}^T\boldsymbol{W}, \boldsymbol{d}_a \in \mathbb{R}^{|V|\times1} \leftarrow {\boldsymbol{e}_a^{\beta}}^T\boldsymbol{W}$ ($\boldsymbol{W}$ is the decoding head of the LLM). Then, the top-N vocabs are predicted for the query and answer: $\boldsymbol{v}_q \leftarrow \textit{top-N}(\boldsymbol{d}_q), \boldsymbol{v}_a \leftarrow \textit{top-N}(\boldsymbol{d}_a)$. If the transformation works, there will be an increased lexical similarity between query and answer. In this place, we use BM25 to compute the similarity score: $\mathrm{BM}_{25}(\boldsymbol{v}_q, \boldsymbol{v}_a)$. As shown in Table \ref{tab:5}, the adaptation (Adapt) exhibits a much higher vocab similarity over the initial LLaMA-2 backbone (Initial). Interestingly, we also observe that the lexical similarity can be improved by fine-tuning as well. In fact, the improved lexical similarity is beneficial to the performance of dense retrieval, as studied by previous works on query and document expansion \cite{nogueira2019doc2query,mao2020generation}. With the unsupervised adaptation, such a transformation is implicitly accomplished for the text embedding. 

$\bullet$ \textbf{Prompt}. The well-adapted LLM encoder is able to make adaptive usage of SELF and NEXT prompt to handle different correlation and paraphrasing relationships (Ada*). Particularly, NEXT-SELF (N2S) is used for correlation, SELF-SELF (S2S) is used for paraphrased documents, and NEXT-NEXT (N2N) is used for paraphrased short texts. The impact of prompt scheme is analyzed with BEIR, which contains retrieval tasks of diversified semantic relationships: for correlation, we have DBPedia (DB), FIQA (FQ), HotpotQA (HQ), NFCorpus (NF), NQ, SCIDOCS (SD), Touch (TO); for long-paraphrasing, we have Arguana (AR); for short paraphrasing, we have Climate-FEVER (CF), FEVER (FV), Quora (QR), SciFact (SF), CQADupStack(CQ). The result is shown in Table \ref{tab:4}, where the prompt utilization exerts a major impact. The majority of tasks are about correlation; thus, N2S works the most effectively in those scenarios. In other paraphrasing cases, S2S (e.g., Arguana) and N2N (e.g., Quora) can help to achieve a better result. 

\begin{table}[t]
    \centering
    \small
    \footnotesize
    \begin{tabular}{p{1.2cm}|C{1.4cm}|C{1.4cm}|C{1.4cm}} 
    \ChangeRT{1pt}
    \textbf{Dim.} & \textbf{DimRed} & \textbf{DimRed}* & \textbf{Sparse} \\  
    \hline
    768  & 41.0 & 41.2 & 41.5 \\
    1024 & 41.0 & 41.3 & 41.9 \\
    2048 & 41.1 & 41.4 & 42.3 \\ 
    4096 & 43.1 & 43.1 & 43.1 \\
    \ChangeRT{1pt}
    \end{tabular}
    \vspace{-5pt}
    \caption{Impact of embedding dimension.} 
    \vspace{-10pt}
    \label{tab:6}
\end{table} 

$\bullet$ \textbf{Dimension}. The LLMs are more expressive than smaller PLMs; however, they also come with higher costs in many perspectives. In this place, we focus on the impact from embedding size, i.e. dimension, which not only affects the computation but also determines the space cost of the vector database. We evaluate alternative dim reduction methods where the embedding size is gradually reduced from 4096 to 768 (Table \ref{tab:6}). One is to jointly learn the LLM and a projection head during fine-tuning (DimRed); another one is to fixed the well fine-tuned LLM retriever and learn the projection head via distillation \cite{liu2022dimension} (DimRed*). Unfortunately, both methods suffer from notably performance loss after dim reduction. We further replace linear projection with sparsification \cite{formal2021splade}, where the top-N entries are selected for the embedding (Sparse). Compared with the first two options, sparsification turns out to be more effective in preserving retrieval performance. The above observations also suggest the necessity of light-weight processing of LLM-based retrievers in the future.

\section{Conclusion}
In this paper, we present Llama2Vec to enhance the LLM-based dense retrieval via unsupervised adaptation. Llama2Vec is composed of two pretext tasks, EBAE and EBAR, where the LLM is prompted to reconstruct the input sentence and predict the following sentence purely with its text embeddings. 
On top of the unsupervised adaptation, the LLM's text embedding can be transformed into a strong representation of global context. By using suitable prompts, it can flexibly support the semantic matching of different correlation and paraphrasing relationships. The effectiveness of Llama2Vec is verified by comprehensive experiments, where the adapted LLM achieves new state-of-the-art performances in both supervised and zero-shot evaluations, indicating the substantial improvements on both accuracy and generalization capability of the retrieval model.



\section{Limitation}
While Llama2Vec has made a substantial progress in adapting the LLM as a strong dense retriever, the current work can still be improved in the following ways. Firstly, the current method is only applied to a 7B model, it remain to explore its impact on larger LLMs. Secondly, the current model is for English centric scenario, it is necessary to make extensions for other languages.  Thirdly, it is also important to find effective ways to maintain an efficient running cost for such large-scale embedding models.


\section{Ethical Consideration}
Llama2Vec is built upon LLaMA-2, it inherits potential biases, toxicity, and other problems present in the underlying LLM. Therefore, we do not recommend utilizing Llama2Vec for retrieval purposes in sensitive contexts. Moreover, the embedding may be influenced by the training data, potentially leading to biased or discriminatory results.

\section{Acknowledgements}
This work is supported by the National Natural Science Foundation of China (Nos. 62272054, 62192784), Beijing Nova Program (No. 20230484319), Xiaomi Young Talents Program, and National Science and Technology Major Project (2023ZD0121504).

\bibliographystyle{acl_natbib}
\bibliography{reference}

\end{document}